\documentclass[11pt]{article}

\usepackage{acl2015}
\usepackage{alltt}
\usepackage{times}
\usepackage{url}
\usepackage{latexsym}

\usepackage[table]{xcolor}
\usepackage{tabularx}
\usepackage{booktabs}
\usepackage{ragged2e}
\usepackage{graphicx}
\usepackage{caption} 
\usepackage{subcaption}
\usepackage{fp}
\usepackage{pgf}
\usepackage{xstring}
\usepackage{collcell}
\usepackage{siunitx}
\usepackage{amsmath}
\usepackage{multirow}

\newcommand{\InHeat}[3]{%
  \IfDecimal{#1}{%
    \pgfmathsetmacro{\IntensityA}{min(1.0, ((min(1.2, #1) - 0.731) / (1.2 - 0.731) + 0.45))}%
    \xdef\IntensityA{\IntensityA}%
    \cellcolor[gray]{\IntensityA}{\num[round-mode=places,round-precision=2]{#1}}%
  }{\textbf{#1}}
}
\newcolumntype{D}{>{\collectcell\InHeat}X<{\endcollectcell}}

\newcommand{\CrossHeat}[3]{%
  \IfDecimal{#1}{%
    \pgfmathsetmacro{\Intensity}{min(1.0, ((#1 - 0.717) / (1.289 - 0.717) + 0.45))}%
    \xdef\Intensity{\Intensity}%
    \cellcolor[gray]{\Intensity}{\num[round-mode=places,round-precision=2]{#1}}%
  }{\textbf{#1}}
}
\newcolumntype{H}{>{\collectcell\CrossHeat}X<{\endcollectcell}}

\newcolumntype{R}[1]{>{\raggedleft\arraybackslash}p{#1}}

\makeatletter

\newcommand{\Rmnum}[1]{\expandafter\@slowromancap\romannumeral #1@}
\makeatother

\title{Generating Sentence Planning Variations for Story Telling}

\author{Stephanie M. Lukin, Lena I. Reed \& Marilyn A. Walker\\
Natural Language and Dialogue Systems\\
University of California, Santa Cruz\\
Baskin School of Engineering\\
\tt{slukin,lireed,mawalker@ucsc.edu}
}
\date{}

\begin{document}
\maketitle
\begin{abstract}

There has been a recent explosion in
applications for dialogue interaction ranging from direction-giving
and tourist information to interactive story systems.  Yet the natural
language generation (NLG) component for many of these systems remains
largely handcrafted.  This limitation greatly restricts the range of
applications; it also means that it is impossible to take advantage of
recent work in expressive and statistical language generation that can
dynamically and automatically produce a large number of variations of
given content. We propose that a
solution to this problem lies in new methods for developing language
generation resources.  
We describe the {\sc es-translator},
a computational language generator
that has previously been applied only to fables, and quantitatively evaluate the domain independence of the {\sc est} by applying it to personal narratives from weblogs.
We then take advantage of recent work on language generation 
to create a parameterized sentence planner for story
generation that provides aggregation operations, variations in discourse and in
point of view. Finally, we present a user evaluation of different
personal narrative retellings.

\end{abstract}


\section{Introduction}
\label{intro-sec}

Recently there has been an explosion in applications for natural
language and dialogue interaction ranging from direction-giving and
tourist information to interactive story systems
\cite{dethlefs2014cluster,Walkeretal11,Huetal15}. While this is due in
part to progress in statistical natural language understanding, many
applications require the system to actually respond in a meaningful
way.  Yet the natural language generation (NLG) component of many
interactive dialogue systems remains largely handcrafted.  This
limitation greatly restricts the range of applications; it also means
that it is impossible to take advantage of recent work in expressive
and statistical language generation that can dynamically and
automatically produce a large number of variations of given content
\cite{RieserLemon11,PaivaEvans2004,Langkilde00,RoweHaLester08,MairesseWalker11}. Such variations are important for expressive 
purposes, we well as for user adaptation and
personalization \cite{ZukermanLitman,Wangetal05,McQuigganetal08}.  We
propose that a solution to this problem lies in new methods for
developing language generation resources.

First we describe the {\sc es-translator} (or {\sc est}), a computational
language generator that has previously been applied only to fables,
e.g. the fable in Table~\ref{fox-crow} \cite{Rishesetal13}. We quantitatively evaluate the domain independence of the {\sc est} by applying it to social
media narratives, such as the {\it Startled Squirrel} story in
Table~\ref{squirrel-blog-story}.
We then present a parameterized
general-purpose framework built on the {\sc est} pipeline, {\sc est 2.0}, that can generate many different tellings of the same story, by utilizing
sentence planning and point of view parameters. Automatically generated story variations are shown in Table~\ref{retell-squirrel-blog-story} and
Table~\ref{retell-fox-crow}.

\begin{table}[t!]
\centering
\begin{small}
\begin{tabular}{|p{2.75in}|}
\hline
\rule{0pt}{2.5ex}  
{\bf Original }   \\ \hline
\rule{0pt}{2.5ex}  
This is one of those times I wish I had a digital camera. We keep a large stainless steel bowl of water outside on the back deck for Benjamin to drink out of when he's playing outside. His bowl has become a very popular site. Throughout the day, many birds drink out of it and bathe in it. 
The birds literally line up on the railing and wait their turn. Squirrels also come to drink out of it. The craziest squirrel just came by- he was literally jumping in fright at what I believe was his own reflection in the bowl. He was startled so much at one point that he leap in the air and fell off the deck. But not quite, I saw his one little paw hanging on! After a moment or two his paw slipped and he tumbled down a few feet. But oh, if you could have seen the look on his startled face and how he jumped back each time he caught his reflection in the bowl! 
\\ \hline
\end{tabular}
\end{small}
\vspace{-0.1in}
\caption{The Startled Squirrel Weblog Story \label{squirrel-blog-story}}
\end{table}

We hypothesize many potential uses for our approach to repurposing and
retelling existing stories.  First, such stories are created daily in
the thousands and cover any topic imaginable. They are natural and
personal, and may be funny, sad, heart-warming or serious.  There are
many potential applications: virtual companions, educational
storytelling, or to share troubles in therapeutic settings
\cite{Bickmore03,PennebakerSeagal99,Gratchetal12}.

Previous research on NLG of linguistic style shows that dialogue
systems are more effective if they can generate stylistic linguistic
variations based on the user's emotional state, personality, style,
confidence, or other factors
\cite{Andreetal00,Piwek03,McQuigganetal08,PorayskaMellish04,ForbesLitman11,Wangetal05,dethlefs2014cluster}.
Other work focuses on variation in journalistic writing or instruction
manuals, where stylistic variations as well as journalistic slant or
connotations have been explored
\cite{Hovy88,GreenDiMarco93,ParisScott94,Poweretal03,InkpenHirst04}.
Previous iterations of the {\sc est} simply presented a sequence of events \cite{Rishesetal13}. This work implements parameterized variation
of linguistic style in the context of weblogs in order to introduce discourse structure into our generated stories.

Our approach differs from previous work on NLG
for narrative because we emphasize (1) domain-independent methods; and
(2) generating a large range of variation, both narratological and
stylistic. \cite{LukinWalker15}'s work on the {\sc est} is the first to generate
dialogue within stories, to have the ability to vary direct
vs. indirect speech, and to generate dialogue utterances
using different stylistic models for character voices.
Previous work can generate
narratological variations, but is domain dependent \cite{CallawayLester02,Montfort07}.

\begin{table}[t!]
\centering
\begin{small}
\begin{tabular}{|p{2.85in}|}
\hline
\rule{0pt}{2.5ex}  
{\bf EST 2.0}  \\ \hline
\rule{0pt}{2.5ex}  
 Benjamin wanted to drink the bowl's water, so I placed the bowl on the deck. The bowl was popular. The birds drank the bowl's water. The birds bathed themselves in the bowl. The birds organized themselves on the deck's railing because the birds wanted to wait. The squirrels drank the bowl's water. The squirrel approached the bowl. The squirrel was startled because the squirrel saw the squirrel's reflection. Because it was startled, the squirrel leapt. The squirrel fell over the deck's railing because the squirrel leaped because the squirrel was startled. The squirrel held the deck's railing with the squirrel's paw. The squirrel's paw slipped off the deck's railing. The squirrel fell.
\\ \hline
\end{tabular}
\end{small}
\vspace{-0.1in}
\caption{Retelling of the Startled Squirrel \label{retell-squirrel-blog-story}}
\end{table}


\begin{table}[t!]
\centering
\begin{small}
\begin{tabular}{|p{2.75in}|}
\hline
\rule{0pt}{2.5ex}  
{\bf Original} \\ \hline
\rule{0pt}{2.5ex}  
A Crow was sitting on a branch of a tree with a piece of cheese in her beak when a Fox observed her and set his wits to work to discover some way of getting the cheese.
Coming and standing under the tree he looked up and said, ``What a noble bird I see above me! Her beauty is without equal, the hue of her plumage exquisite. If only her voice is as sweet as her looks are fair, she ought without doubt to be Queen of the Birds.''
The Crow was hugely flattered by this, and just to show the Fox that she could sing she gave a loud caw.
Down came the cheese,of course, and the Fox, snatching it up, said, ``You have a voice, madam, I see: what you want is wits.'' \\
\hline 
\end{tabular}
\end{small}
\vspace{-0.1in}
\caption{``The Fox and the Crow'' \label{fox-crow}}
\end{table}

Sec.~\ref{corpus-method-sec} describes our corpus of stories and the
architecture of our story generation framework, {\sc est 2.0}.\footnote{The corpus is
  available from \url{https://nlds.soe.ucsc.edu/personabank}.}
  Sec.~\ref{blog-sec} describes experiments testing the coverage and
  correctness of {\sc est 2.0}.  Sec.~\ref{exp-sec} describes
  experiments testing user perceptions of different linguistic
  variations in storytelling.  Our contributions are:
\vspace{-0.05in}
\begin{itemize} 
\item We produce {\sc sig} representations of 100
personal narratives from a weblog corpus, using the story annotation tool
Scheherezade \cite{ElsonMcKeown09,Elson-Thesis12};  
\vspace{-0.1in}
\item We compare {\sc est 2.0} to {\sc est} and show how we have not only made improvements to the translation algorithm, but can extend and compare to personal narratives.
\vspace{-0.1in}
\item We implement a parameterized variation
of linguistic style in order to introduce discourse structure into our generated narratives.
\vspace{-0.1in}
\item We carry out
experiments to gather user perceptions of different
sentence planning choices that can be made with complex sentences in
stories.  
\end{itemize}

We sum up and discuss future work in Sec.~\ref{conc-sec}.

\section{Story Generation Framework}
\label{corpus-method-sec} 

\begin{figure}[thb!]
\begin{center}
\includegraphics[width=3.in]{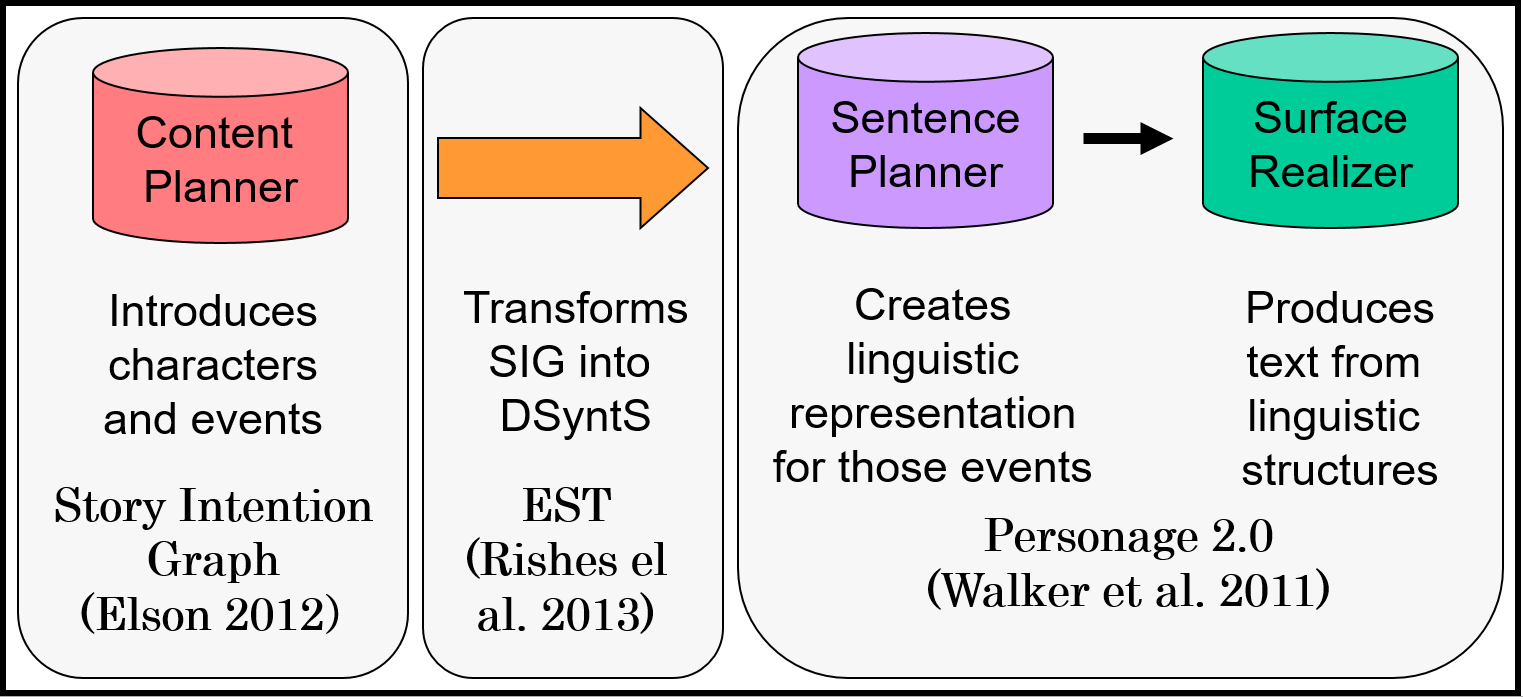}
\caption{\label{est-arch-fig} {NLG pipeline method of the ES Translator.}}
\end{center}
\vspace{-0.2in}
\end{figure}


Fig.~\ref{est-arch-fig} illustrates our overall architecture,
which uses NLG modules to separate the process of planning {\it What to
  say} (content planning and selection, fabula) from decisions about
{\it How to say it} (sentence planning and realization, discourse).
We build on three existing tools from previous work: the {\sc
  scheherezade} story annotation tool, the {\sc personage} generator,
and the {\sc es-translator} ({\sc est})
\cite{Elson-Thesis12,MairesseWalker11,Rishesetal13}.  The {\sc est}
uses the {\sc story intention graph} ({\sc sig}) representation
produced by {\sc scheherezade} and its theoretical grounding as a
basis for the content for generation.  The {\sc est} bridges the
narrative representation of the {\sc sig} to the representation
required by {\sc personage} by generating the text plans and the deep
syntactic structures that {\sc personage} requires. Thus any story or
content represented as a {\sc sig} can be retold using {\sc
  personage}. See Fig.~\ref{est-arch-fig}.

There are several advantages to using the {\sc sig} as the representation for a content pool:

\begin{itemize}
\item  Elson's {\sc dramabank} provides stories encoded as {\sc sig}s including 36 Aesop's Fables, such as  {\it The Fox and the Crow}
in Table~\ref{fox-crow}.
  \vspace{-0.1in}
\item The {\sc sig} framework includes an annotation tool called
  {\sc scheherazade} that supports representing any 
narrative as a  {\sc sig}.
  \vspace{-0.1in}
\item {\sc scheherezade} comes with a realizer that regenerates
stories from the {\sc sig}: this realizer provides alternative
story realizations that we can compare to the {\sc est 2.0} output.
\end{itemize}
\vspace{-0.05in}

We currently have 100 personal narratives annotated with the {\sc sig}
representation on topics such as travel, storms, gardening, funerals,
going to the doctor, camping, and snorkeling, selected from a corpus
of a million stories \cite{GordonSwanson09}. We use the stories in
Tables~\ref{squirrel-blog-story} and~\ref{fox-crow} in this
paper to explain our framework. 

\begin{figure}[tb]
\begin{center}
\includegraphics[width=3.0in]{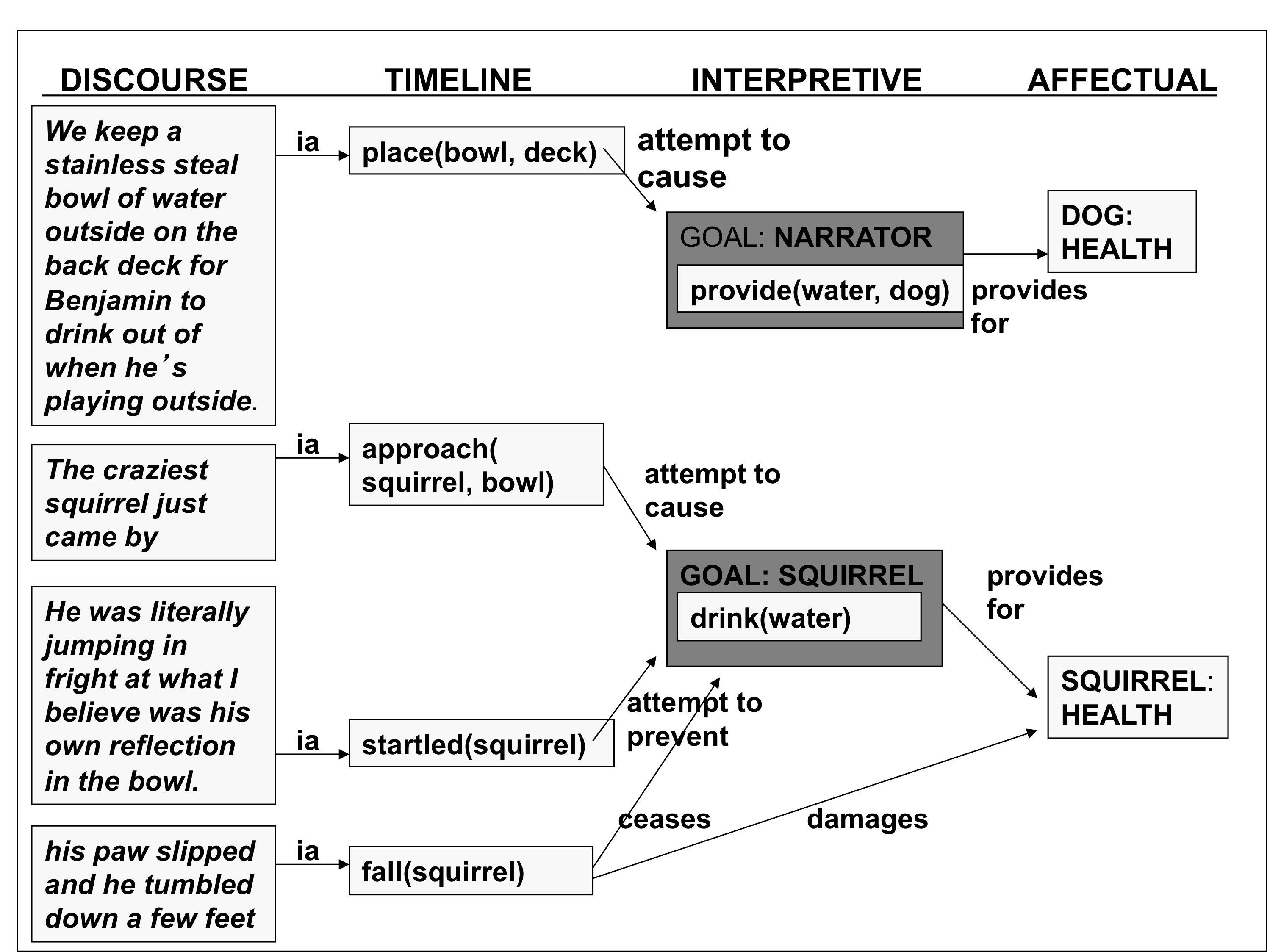}
\caption{\label{squirrel-sig} Part of the {\sc Story Intention Graph} ({\sc sig}) for {\it The Startled Squirrel}.}
\vspace{-0.25in}
\end{center}
\end{figure}

Fig.~\ref{squirrel-sig} shows the {\sc sig} for {\it The Startled
  Squirrel} story in Table~\ref{squirrel-blog-story}.  To create a
{\sc sig}, {\sc Scheherazade} annotators: (1) identify key
entities; (2) model events and statives as propositions and arrange
them in a timeline; and (3) model the annotator's understanding of the
overarching goals, plans and beliefs of the story's agents. {\sc
  scheherazade} allows users to annotate a story along several
dimensions, starting with the surface form of the story (first column
in Table~\ref{squirrel-sig}) and then proceeding to deeper
representations. The first dimension (second column in
Table~\ref{squirrel-sig}) is called the ``timeline layer'', in which
the story is encoded as predicate-argument structures (propositions)
that are temporally ordered on a timeline.  {\sc scheherazade} adapts
information about predicate-argument structures from the VerbNet
lexical database \cite{Kipperetal06} and uses WordNet
\cite{Fellbaum98} as its noun and adjectives taxonomy. The arcs of the
story graph are labeled with discourse relations, such as {\it attempts
to cause}, or {\it temporal order} (see Chapter 4 of \cite{Elson-Thesis12}.)

\begin{table}[t!]
\centering
\begin{small}
\begin{tabular}{|p{2.75in}|}
\hline
\rule{0pt}{2.5ex}  
{\bf EST 2.0} \rule{0pt}{2.5ex}   \\ \hline
\rule{0pt}{2.5ex}  
The crow sat on the tree's branch. The cheese was in the crow's pecker. The crow thought ``I will eat the cheese on the branch of the tree because the clarity of the sky is so-somewhat beautiful." The fox observed the crow. The fox thought ``I will obtain the cheese from the crow's nib." The fox came.
The fox stood under the tree. The fox looked toward the crow. The fox avered ``I see you!" The fox alleged `your's beauty is quite incomparable, okay?" The fox alleged `your's feather's chromaticity is damn exquisite." The fox said ``if your's voice's pleasantness is equal to your's visual aspect's loveliness you undoubtedly are every every birds's queen!" The crow thought ``the fox was so-somewhat flattering." The crow thought ``I will demonstrate my voice." The crow loudly cawed.
The cheese fell. The fox snatched the cheese. The fox said ``you are somewhat able to sing, alright?"
The fox alleged ``you need the wits!"\\ \hline
\end{tabular}
\end{small}
\vspace{-0.1in}
\caption{Retelling of ``The Fox and the Crow'' \label{retell-fox-crow}}
\vspace{-0.2in}
\end{table}

\begin{table*}[]
\centering
\begin{small}
\begin{tabular}{|r|p{2.4in}|p{2.4in}|}
\hline
\bf Variation & \bf Blog Output & \bf Fable Output  \\ \hline
\rule{0pt}{2.5ex}  
\bf Original & We keep a large stainless steel bowl of water outside on the back deck for Benjamin to drink out of when he's playing outside. 
& The Crow was hugely flattered by this, and just to show the Fox that she could sing she gave a loud caw.
\\ \hline 
\rule{0pt}{2.5ex}  
\bf Sch & A narrator placed a steely and large bowl on a back deck in order for a dog to drink the water of the bowl. 
& The crow cawed loudly in order for she to show him that she was able to sing.
\\ \hline
\rule{0pt}{2.5ex}  
\bf EST 1.0 & I placed the bowl on the deck in order for Benjamin to drink the bowl's water. 
& The crow cawed loudly in order to show the fox the crow was able to sing.
\\ \hline
\rule{0pt}{2.5ex}  
\bf  becauseNS & I placed the bowl on the deck because Benjamin wanted to drink the bowl's water. 
& The crow cawed loudly because she wanted to show the fox the crow was able to sing.
\\ \hline
\rule{0pt}{2.5ex}  
\bf becauseSN  & Because Benjamin wanted to drink the bowl's water, I placed the bowl on the deck. 
& Because the crow wanted to show the fox the crow was able to sing, she cawed loudly.
\\ \hline
\rule{0pt}{2.5ex}  
\bf NS  & I placed the bowl on the deck. Benjamin wanted to drink the bowl's water. 
& The crow cawed loudly. She wanted to show the fox the crow was able to sing. 
\\ \hline
\rule{0pt}{2.5ex}  
\bf N  & I placed the bowl on the deck. 
& The crow cawed loudly.
\\ \hline
\rule{0pt}{2.5ex}  
\bf soSN  & Benjamin wanted to drink the bowl's water, so I placed the bowl on the deck. 
& The crow wanted to show the fox the crow was able to sing, so she cawed loudly. 
\\ \hline
\end{tabular}
\caption{Sentence Planning Variations added to
 {\sc est 2.0} for Contingency relations, exemplified by 
{\it The Startled Squirrel} and {\it The Fox and the Crow}. 
Variation {\bf N} is intended to test whether the content of the satellite
can be recovered from context. {\bf Sch} is the realization produced
by Scheherezade. \label{nv-v1-fig}}
\end{small}
\end{table*}

The {\sc est} applies a model of syntax to the {\sc sig} which
translates from the semantic representation of the {\sc sig} to the
syntactic formalism of Deep Syntactic Structures ({\sc dsynts})
required by the {\sc personage} generator
\cite{LavoieRambow97,Melcuk88,MairesseWalker11}. 
Fig.~\ref{est-arch-fig} provides a high level view of the
architecture of {\sc est}. The full translation methodology is
described in \cite{Rishesetal13}.

{\sc dsynts} are a flexible dependency tree representation of an
utterance that gives us access to the underlying linguistic structure
of a sentence that goes beyond surface string manipulation. The nodes
of the {\sc dsynts} syntactic trees are labeled with lexemes and the
arcs of the tree are labeled with syntactic relations. The {\sc
  dsynts} formalism distinguishes between arguments and modifiers and
between different types of arguments (subject, direct and indirect
object etc). Lexicalized nodes also contain a range of grammatical
features used in generation. RealPro handles morphology, agreement and
function words to produce an output string.

This paper utilizes the ability of the {\sc est 2.0} and the flexibility of {\sc dsynts} to produce direct speech that varies the character voice as illustrated in Table~\ref{retell-fox-crow} \cite{LukinWalker15}. 
By simply modifying the
{\small \tt person} parameter in the {\sc dsynts}, we can change the
sentence to be realized in the first person. For example, to produce
the variations in Table~\ref{retell-fox-crow}, we use 
both first person, and direct speech, as well as linguistic
styles from {\sc personage}: a neutral voice
for the narrator, a shy voice for the crow, and a laid-back 
voice for the fox \cite{LukinWalker15}. We fully utilize this variation when we retell personal narratives in  {\sc est 2.0}. 

\begin{table}[ht!]
\centering
\begin{scriptsize}
\caption{1: original unbroken {\sc dsynts}; 2) deaggregated {\sc dsynts}; 3) contingency text plan\label{deagg-ex}}
\vspace{-0.1in}
\begin{tabular}{|p{3.0in}|}
\hline 
\vspace{0.057in}
{\bf 1: ORIGINAL}
\begingroup
\fontsize{7pt}{7pt}
\begin{verbatim}
<dsynts id="5_6">
 <dsyntnode class="verb" lexeme="organize" 
   mode="" mood="ind" rel="II" tense="past">
  <dsyntnode article="def" class="common_noun" 
     lexeme="bird" number="pl" person="" rel="I"/>
  <dsyntnode article="def" class="common_noun" 
     lexeme="bird" number="pl" person="" rel="II"/>
  <dsyntnode class="preposition" lexeme="on" 
          rel="ATTR">
   <dsyntnode article="def" class="common_noun" 
     lexeme="railing" number="sg" person="" rel="II">
     <dsyntnode article="no-art" class="common_noun" 
     lexeme="deck" number="sg" person="" rel="I"/>
   </dsyntnode>
  </dsyntnode>
  <dsyntnode class="preposition" lexeme="in_order" 
       rel="ATTR">
    <dsyntnode class="verb" extrapo="+" lexeme="wait" 
       mode="inf-to" mood="inf-to" 
       rel="II" tense="inf-to">
      <dsyntnode article="def" class="common_noun" 
       lexeme="bird" number="pl" person="" rel="I"/>
    </dsyntnode>
  </dsyntnode>
</dsyntnode>
</dsynts>
\end{verbatim} 
\endgroup
\\ \hline
\vspace{0.057in}
{\bf 2: DEAGGREGATION}
\begingroup
\fontsize{7pt}{7pt}
\begin{alltt}
<dsynts id="5">
  <dsyntnode class="verb" lexeme="organize" 
     mood="ind" rel="II" tense="past">
    <dsyntnode article="def" class="common_noun" 
     lexeme="bird" number="pl" person="" rel="I"/>
    <dsyntnode article="def" class="common_noun" 
     lexeme="bird" number="pl" person="" rel="II"/>
    <dsyntnode class="preposition" lexeme="on" 
               rel="ATTR">
     <dsyntnode article="def" class="common_noun" l
       lexeme="railing" number="sg" 
       person="" rel="II">
      <dsyntnode article="no-art" class="common_noun" 
       lexeme="deck" number="sg" person="" rel="I"/>
      </dsyntnode>
    </dsyntnode>
  </dsyntnode>
</dsynts>

<dsynts id="6">
  <dsyntnode class="verb" lexeme="want"
        mood="ind" rel="II" tense="past">
    <dsyntnode article="def" class="common_noun" 
        lexeme="bird" number="pl" person="" r
    <dsyntnode class="verb" extrapo="+" 
     lexeme="wait" mode="inf-to" mood="inf-to" 
     rel="II" tense="inf-to"/>
  </dsyntnode>
</dsynts>
\end{alltt} 
\endgroup
\\ \hline
\vspace{0.057in}
{\bf 3: AGGREGATION TEXT PLAN}
\begingroup
\fontsize{7pt}{7pt}
\begin{alltt}
  <speechplan voice="Narrator">
  <rstplan>
    <relation name="contingency_cause">
      <proposition id="1" ns="nucleus"/>
      <proposition id="2" ns="satellite"/>
    </relation>
  </rstplan>
  <proposition dialogue_act="5" id="1"/>
  <proposition dialogue_act="6" id="2"/>
</speechplan>
\end{alltt} 
\endgroup
\\ \hline
\end{tabular}
\end{scriptsize}
\vspace{-0.1in}
\end{table}

This paper and introduces support for new discourse relations, such as
aggregating clauses related by the contingency discourse relation (one
of many listed in the Penn Discourse Tree Bank (PDTB)
\cite{prasad2008penn}). In {\sc sig} encoding, contingency clauses are
always expressed with the ``in order to'' relation
(Table~\ref{deagg-ex}, 1). To support linguistic variation, we introduce ``de-aggregation'' onto these aggregating
clauses in order to have the flexibility to rephrase, restructure, or
ignore clauses as indicated by our parameterized sentence planner. We
identify candidate story points in the {\sc sig} that contain a
contingency relation (annotated in the Timeline layer) and
deliberately break apart this hard relationship to create nucleus and
satellite {\sc dsynts} that represents the entire sentence  
(Table~\ref{deagg-ex}, 2) \cite{mann1988rhetorical}. We create a text plan
(Table~\ref{deagg-ex}, 3) to allow the sentence planner to
reconstruct this content in various ways. Table~\ref{nv-v1-fig} shows
sentence planning variations for the contingency relation for both
fables and personal narratives ({\bf soSN}, {\bf becauseNS}, {\bf
  becauseSN}, {\bf NS}, {\bf N}), the output of {\sc est 1.0}, the
original sentence ({\bf original}), and the {\sc scheherazade}
realization ({\bf Sch}) which provides an additional baseline.  The
{\bf Sch} variant is the original ``in order to'' contingency
relationship produced by the {\sc sig} annotation.  The {\bf
  becauseNS} operation presents the {\it nucleus} first, followed by a
{\it because}, and then the {\it satellite}. We can also treat the
nucleus and satellite as two different sentences ({\bf NS}) or
completely leave off the satellite ({\bf N}). We believe the {\bf N}
variant is useful if the satellite can be easily inferred from the
prior context.

The richness of the discourse information present in the {\sc sig} and
our ability to de-aggregate and aggregate will enable us to implement
other discourse relations in future work.

\section{Personal Narrative Evaluation}
\label{blog-sec}
\vspace{-0.1in}

After annotating our 100 stories with the {\sc scheherazade}
annotation tool, we ran them through the {\sc est}, and examined the
output. We discovered several bugs arising from variation in the blogs
that are not present in the Fables, and fixed them.  In previous work
on the {\sc est}, the machine translation metrics Levenshtein's
distance and BLEU score were used to compare the original Aesop's
Fables to their generated {\sc est} and {\sc scheherazade}
reproductions (denoted {\bf EST} and {\bf Sch})
\cite{Rishesetal13}. These metrics are not ideal for evaluating story
quality, especially when generating stylistic variations of the
original story. However they allow us to automatically test some
aspects of system coverage, so we repeat this evaluation on the blog
dataset.  

Table~\ref{bleu-fables} presents BLEU and Levenshtein scores for the
original 36 Fables and all 100 blog stories, compared to both {\bf
  Sch} and {\sc est 1.0}.  Levenshtein distance computes the minimum
edit distance between two strings, so we compare the entire original
story to a generated version. A lower score indicates a closer
comparison. BLEU score computes the overlap between two strings taking
word order into consideration: a higher BLEU score indicates a closer
match between candidate strings.  Thus Table~\ref{bleu-fables}
provides quantitative evidence that the style of the original blogs is
very different from Aesop's Fables. Neither the {\sc est} output nor
the {\bf Sch} output comes close to representing the original textual
style (Blogs Original-Sch and Original-EST).

\begin{table}[h!]
\begin{center}
\caption{Mean for Levenshtein and BLEU on the Fables development set vs. the Blogs \label{bleu-fables} }
\vspace{-.1in}
\begin{tabular}{|l|r |c|c|}
\hline
 & & \bf Lev  & \bf BLEU  \\ \hline \hline
\bf FABLES & Sch-EST 		& 72  & .32  \\ \hline
& Original-Sch		& 116  & .06  \\ \hline
& Original-EST		& 108  & .03 \\ \hline

\bf BLOGS &  Sch-EST 		& 110  & .66   \\ \hline
&  Original-Sch	& 736  & .21   \\ \hline
&  Original-EST		& 33  & .21  \\ \hline
\end{tabular}
\end{center}
\end{table}

However we find that {\bf EST} compares favorably to {\bf Sch} on the
blogs with a relatively low Levenshtein score, and higher BLEU score 
(Blogs Sch-EST) than the original Fables evaluation (Fables
Sch-EST). This indicates that even though the blogs have a diversity
of language and style, our translation
comes close to the {\bf Sch} baseline.

\section{Experimental Design and Results}
\label{exp-sec}
\label{results-sec}

We conduct two experiments on Mechanical Turk to test variations generated
with the de-aggregation and point of view parameters. We compare the
variations amongst themselves and to the original sentence in a story.
We are also interested in identifying differences among individual
stories.

In the first experiment, we show an excerpt from the original story telling and indicate to the participants that ``any of the following sentences could come next in the story''. We then list all variations of the following 
sentence with the ``in order to'' contingency relationship (examples from the {\it Startled Squirrel} labeled {\sc est 2.0} in Table~\ref{nv-v1-fig}).

Our aim is to elicit rating of the variations
in terms of correctness and goodness of fit within the story context
(1 is best, 5 is worst), and to rank the sentences by personal
preference (in experiment 1 we showed 7 variations where 1 is best, 7 is worst; in experiment 2 we showed 3 variations where 1
is best, 3 is worst). We also show the original blog sentence and the EST
1.0 output before de-aggregation and sentence planning. We
emphasize that the readers should read each variation {\it in the
  context of the entire story} and encourage them to reread the story
with each new sentence to understand this context.


In the second experiment, we compare the 
original sentence with our best realization, and the realization
produced by {\sc scheherezade} ({\bf Sch}).
We expect  that {\sc scheherezade} will score more
poorly in this instance because it cannot
change point of view from third person to first person,
even though its output is more fluent than {\sc est 2.0} for many
cases.

\subsection{Results Experiment 1}
We had 7 participants analyze each of the 16 story segments. All participants were native English speakers. Table~\ref{exp1-stats} shows the means and standard deviations for
correctness and preference rankings in the first experiment. 
We find that averaged across all stories, there is a clear order for correctness and preference: original, soSN, becauseNS, becauseSN, NS, EST, N. 

We performed an ANOVA on preference and found that story has no
significant effect on the results (F(1, 15) $=$ 0.18, p $=$ 1.00),
indicating that all stories are well-formed and there are no
outliers in the story selection. On the other hand, realization does
have a significant effect on preference (F(1, 6) $=$ 33.74, p $=$
0.00). This supports our hypothesis that the realizations are distinct
from each other and there are preferences amongst them.



\begin{figure}
\begin{center}
\includegraphics[width=3.0in]{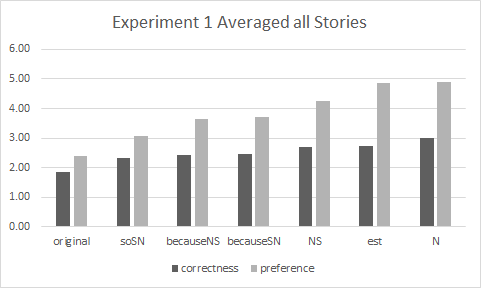}
\caption{\label{hist-all-exp1} Histogram of Correctness and Preference for Experiment 1 averaged across story (lower is better)}
\end{center}
\end{figure}

Fig.~\ref{hist-all-exp1} shows the average correctness and preference
for all stories. Paired t-tests show that
there is a significant difference in reported
correctness between {\bf orig} and {\bf soSN} (p $<$ 0.05), but no
difference between {\bf soSN} and {\bf becauseNS} (p = 0.133), or {\bf
  becauseSN} (p = 0.08).
There is a difference between {\bf soSN} and {\bf NS} (p $<$ 0.005),
as well as between the two different {\bf because} operations and {\bf
  NS} (p $<$ 0.05). There are no other significant differences.

\begin{table*}[hbt!]
\begin{center}
\begin{tabular}{|lr|c|c|c|c|c|c|c|}
\hline
& & \bf Orig & \bf soSN & \bf becauseNS & \bf becauseSN & \bf NS & \bf EST & \bf N \\ \hline \hline
\bf ALL & \bf C & 1.8  & 2.3  & 2.4  & 2.5  & 2.7  & 2.7  & 3.0 \\ \hline
 & \bf P & 2.4  & 3.1  & 3.7 & 3.8  & 4.2  & 4.9  & 4.9  \\ \hline \hline

\bf Protest & \bf C  & 4.9  & 2.7  & 2.4 & 3.9  & 2.1  & 2.7 & 2.7  \\ \hline
 & \bf   P & 1.0  & 4.1 & 4.3  & 4.4  & 4.4  & 4.4  & {\bf 2.8 } \\ \hline \hline
\bf Story 042 & \bf C & 4.2  & 4.2  & 4.3 & 3.8 & 3.7 & 4.2 & 2.7  \\ \hline
 & \bf P & 3.3  & 3.7  & 3.6  & 4.6 & {\bf 3.1} & 5  & 4 \\ \hline
\end{tabular}
\caption{Exp 1: Means for correctness {\bf C} and preference {\bf P} for original sentences and generated variations for {\bf ALL} stories vs. 
the {\bf Protest Story} and {\bf a042} (stimuli in Table~\ref{story-table}).
Lower is better.
\label{exp1-protest-stats} \label{exp1-stats} }
\end{center}
\end{table*}


The are larger differences on the preference metric.
Paired t-tests show that there is a
significant difference between {\bf orig} and {\bf soSN} (p $<$ 0.0001) and {\bf soSN} and {\bf becauseNS}
(p $<$ 0.05). There is no difference in
preference between {\bf becauseNS} and {\bf becauseSN} (p =
0.31). However there is a significant difference between {\bf soSN}
and {\bf becauseSN} (p $<$ 0.005) and {\bf becauseNS} and {\bf NS} (p
$<$ 0.0001). Finally, there is significant difference between {\bf
  becauseSN} and {\bf NS} (p $<$ 0.005) and {\bf NS} and {\bf
  EST} (p $<$ 0.005). There is no difference between {\bf
  EST} and {\bf N} (p = 0.375), but there is a difference between
{\bf NS} and {\bf N} (p $<$ 0.05).

These results indicate that the original sentence, as expected, is the
most correct and preferred. Qualitative feedback 
on the original sentence included: ``The
one I ranked first makes a more interesting story. Most of the others
would be sufficient, but boring.''; ``The sentence I ranked first
makes more sense in the context of the story. The others tell you
similar info, but do not really fit.''. Some participants ranked {\bf
  soSN} as their preferred variant (although the difference was never
statistically significant): ``The one I rated the best sounded really
natural.''

Although we observe an overall ranking trend, there are some
differences by story for {\bf NS} and {\bf N}. Most of the time, these
two are ranked the lowest. Some subjects observe: ``\#1 [{\bf orig}]
\& \#2 [{\bf soSN}] had a lot of detail.  \#7 [{\bf N}] did not
explain what the person wanted to see'' (a044 in Table~\ref{story-table}); 
``The sentence I rated the
worst [{\bf N}] didn't explain why the person wanted to cook them, but
it would have been an okay sentence.'' (a060 in Table~\ref{story-table}); 
``I ranked the lower number [{\bf N}] because they either did not contain the full thought of the
subject or they added details that are to be assumed.'' (a044 in Table~\ref{story-table}); 
``They were
all fairly good sentences. The one I ranked worst [{\bf N}] just left
out why they decided to use facebook.'' (a042 in Table~\ref{story-table}).

However, there is some support for {\bf NS} and {\bf N}. We also find that there is a significant interaction between story and realization (F(2, 89) $=$ 1.70, p $=$ 0.00), thus subjects' preference of the realization are based on the story they are reading. One subject
commented: ``\#1 [{\bf orig}] was the most descriptive about what
family the person is looking for.  I did like the way \#3 [{\bf NS}]
was two sentences.  It seemed to put a different emphasis on finding
family'' (a042 in Table~\ref{story-table}). Another thought that the explanatory utterance altered the
tone of the story: ``The parent and the children in the story were
having a good time. It doesn't make sense that parent would want to do
something to annoy them [the satellite utterance]'' (a060 in Table~\ref{story-table}). This person
preferred leaving off the satellite and ranked {\bf N} as the highest
preference.


We examined these interactions between story and preference ranking
for {\bf NS} and {\bf N}.  This may be depend on either context or on
the {\sc sig} annotations. For example, in one story (protest in Table~\ref{story-table}) 
our best
realization {\bf soSN}, produces: ``The protesters wanted to block the
street, so the person said for the protesters to protest in the street
in order to block it.'' and {\bf N} produces ``The person said for the
protesters to protest in the street in order to block it.''. One
subject, who ranked {\bf N} second only to {\bf original}, observed:
``Since the police were coming there with tear gas, it appears the
protesters had already shut things down. There is no need to tell them
to block the street.'' Another subject who ranked {\bf N} as second
preference similarly observed ``Frankly using the word protesters and
protest too many times made it seem like a word puzzle or riddle. The
meaning was lost in too many variations of the word `protest.' If the
wording was awkward, I tried to assign it toward the `worst' end of
the scale. If it seemed to flow more naturally, as a story would, I
tried to assign it toward the `best' end.'' 

\begin{figure}[h!]
\begin{center}
\includegraphics[width=3.0in]{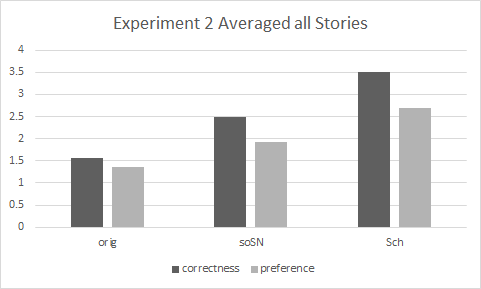}
\caption{\label{hist-all-exp2} Histogram of Correctness and Preference for Experiment 1 averaged across story (lower is better)}
\end{center}
\vspace{-0.15in}
\end{figure}

Although the means
in this story seem very distinct
(Table~\ref{exp1-protest-stats}), there is only a significant
difference between {\bf orig} and {\bf N} (p $<$ 0.005) and {\bf N}
and {\bf EST} (p $<$ 0.05).
Table~\ref{exp1-protest-stats} also includes  the means 
for story a042 (Table~\ref{story-table}) where {\bf NS} is ranked
highest for preference. Despite this, the only significant difference
between {\bf NS} is with {\bf EST 1.0} (p $<$ 0.05).

\subsection{Results Experiment 2} 

Experiment 2 compares our best realization to the {\sc scheherazade}
realizer, exploiting the ability of {\sc est 2.0} to change the point
of view.  Seven participants analyzed each of the 16 story
segments. All participants were native English speakers.


\begin{table}[h!tb]
\begin{center}
\begin{tabular}{|r|c|c|c|}
\hline
 & \bf Original & \bf soSN & \bf Sch \\ \hline \hline
\bf Correctness & 1.6  & 2.5  & 3.5  \\ \hline
\bf Preference & 1.4  & 1.9  & 2.7 \\ \hline

\end{tabular}
\caption{Exp 2: Means for correctness and preference for original sentence, our best realization {\bf soSN}, and Sch. Lower is better.
\label{exp2-stats}}
\end{center}
\vspace{-0.15in}
\end{table}


Table~\ref{exp2-stats} shows the means for
correctness and preference rankings.
Figure~\ref{hist-all-exp2} shows a histogram of average correctness
and preference by realization for all stories. 
There is a clear order for correctness and
preference: original, soSN, Sch, with significant
differences between all pairs of realizations (p $<$ 0.0001).


However, in six of the 19 stories, there is
no significant difference between Sch and soSN.
Three of them do not contain ``I'' or ``the
narrator" in the realization sentence. Many of the subjects comment
that the realization with ``the narrator'' does not follow the style
of the story: ``The second [{\bf Sch}] uses that awful `narrator.''' (a001 in Table~\ref{story-table});
``Forget the narrator sentence. From here on out it's always the
worst!'' (a001 in Table~\ref{story-table}). We hypothesize that in the three sentences without ``the
narrator'', {\bf Sch} can be properly evaluated without the
``narrator'' bias. In fact, in these situations, {\bf Sch} was rated
higher than {\bf soSN}: ``I chose the sentences in order of best
explanatory detail'' ({\it Startled Squirrel} in Table~\ref{nv-v1-fig}).

Compare the {\bf soSN} realization in the protest story in Table~\ref{story-table} ``The leaders wanted to talk, so
they met near the workplace.'' with {\bf Sch} ``The group of leaders
was meeting in order to talk about running a group of countries and
near a workplace.'' {\bf Sch} has so much more detail than {\bf
  soSN}. While the {\sc EST} has massively improved and overall is preferred
to {\bf Sch}, some semantic components are 
lost in the translation process.


\section{Discussion and Conclusions}
\label{conc-sec} 

To our knowledge, this is the first time that sentence planning
variations for story telling have been implemented in a framework
where the discourse (telling) is completely independent of the fabula
(content) of the story \cite{Lonneker05}. We also show for the first
time that the {\sc scheherezade} annotation tool can be applied to informal narratives
such as personal narratives from weblogs, and the resulting
{\sc sig} representations work with existing tools
for translating from the {\sc sig} to a retelling of a story.

We present a parameterized sentence planner for story generation, that
provides aggregation operations and variations in point of view.  The
technical aspects of de-aggregation and aggregation 
builds on previous work in NLG and our earlier work on SPaRKy
\cite{cahilletalacl01,ScottSouza90,ParisScott94,NakatsuWhite10,Howcroftetal13,Walkeretal07,StentMolina09}.
However we are not aware of previous NLG applications needing to first
de-aggregate the content, before applying aggregation
operations.

Our experiments show that, as expected, readers almost always prefer
the original sentence over automatically produced variations, but that
the {\bf soSN} variant is preferred. We examine two specific stories
where preferences vary from the overall trend: these stories suggest
future possible experiments where we might vary more aspects of the
story context and audience. We also compare our best variation to what
{\sc scheherazade} produces. Despite the fact that the {\sc
  scheherazade} realizer was targeted at the {\sc sig}, our best
variant is most often ranked as a preferred choice.

In future work, we aim to explore interactions between a number of our
novel narratological parameters.  We expect to do this both with a 
rule-based approach, as well as by building on recent work on
statistical models for expressive generation
\cite{RieserLemon11,PaivaEvans2004,Langkilde00,RoweHaLester08,MairesseWalker11}. This
should allow us to train a narrative generator to achieve particular
narrative effects, such as engagement or empathy with particular
characters. We will also expand the discourse relations that
{\sc est 2.0} can handle.

\vspace{0.1in}
\noindent{\bf Acknowledgements.} This research was supported by Nuance Foundation Grant SC-14-74,
NSF Grants IIS-HCC-1115742 and IIS-1002921.

\vspace{0.1in}
\noindent{\bf Appendix.}
Table~\ref{story-table} provides additional examples of the output
of the {\sc est 2.0} system, illustrating particular user preferences and system
strengths and weaknesses.

\begin{table*}
\centering
\begin{small}
\begin{tabular}{|r|p{5.0in}|}
\hline
a001 & Bug out for blood the other night, I left the patio door open just long enough to let in a dozen bugs of various size. I didn't notice them until the middle of the night, when I saw them clinging to the ceiling. 
I grabbed the closest object within reach, and with a rolled-up comic book I smote mine enemies and smeared their greasy bug guts. All except for the biggest one. 
I only clipped that one, taking off one of its limbs. But it got away before I could finish the job. So now there's a five-limbed insect lurking in the apartment, no doubt looking for some vengeance against me. \\ \hline 
orig & I'm looking around corners, checking the toilet before sitting down \\
Sch & The narrator began to check the toilet seat of the narrator for the leader of the group of bugs in order for she to sit down on the toilet seat of the narrator	\\
EST & I wanted to sit down on my toilet seat, so I checked my toilet seat for the bugs's leader.\\
\hline \hline
a042-1 &  This last week I was exploring the FamilySearch Learning Center area to see what was new. I randomly choose a video to check it out. In the middle of the video he was talking about using Facebook as a research tool. I listened as he talked about finding a family group on Facebook, on which he found pictures and information. \\ \hline 
orig & I considered this and decided it was time to set up our Family Facebook to find those branches that have not yet been found. \\
Sch & The narrator decided to use the utility in order to find the family of the narrator. \\
soSN & I wanted to find my family, so I decided to use Facebook.\\
becauseNS & I decided to use Facebook because I wanted to find my family.\\
becauseSN & Because I wanted to find my family, I decided to use Facebook.\\
NS & I decided to use Facebook. I wanted to find my family.\\ 
EST & I decided to use Facebook in order for me to find my family.\\
N & I decided to use Facebook. \\
\hline \hline
a044 & I'm writing this from the Bogota airport, waiting for my flight back to Santiago. When I last posted, I was about to head off to northern Colombia to Cartagena. So, how was it? 	\\ \hline 
orig & I split the 4 days between Cartagena and Santa Marta \\
Sch & \\
soSN & I wanted to see Cartagena, so I traveled to Colombia.	\\
becauseNS & I traveled to Colombia because I wanted to see Cartagena.	\\
becauseSN & Because I wanted to see Cartagena, I traveled to Colombia.	\\
NS & I traveled to Colombia. I wanted to see Cartagena.	 \\
EST & I traveled to Colombia in order for me to see Cartagena and for me to see Santa Marta.\\
N & I traveled to Colombia.	\\
\hline \hline
a060-1 & I hope everyone survived the snow! With the early school dismissal on Friday, it felt like a 3 day weekend. 
My kids are just not creatures of Winter. I did manage to take them and some of the neighborhood kids out sledding on Friday and Saturday. That was a blast. 
The kids had more fun, and I had a fire in the shelter with a bag of marshmallows and just enjoying myself. Followed up, of course, with hot chocolate at home. 
I even managed to cook cornbread from scratch, in an old (my grandmothers) cast-iron skillet, with chicken and gravy for dinner. \\ \hline 
orig & If I had any collard greens, I think I would have cooked them too (just to annoy the kids).\\
Sch & The narrator wanted to cook a group of collards in order to annoy the group of children of the narrator. \\
soSN & I wanted to annoy my children, so I wanted to cook the collards.\\
becauseNS & I wanted to cook the collards because I wanted to annoy my children. \\
becauseSN & Because I wanted to annoy my children, I wanted to cook the collards.\\
NS & I wanted to cook the collards. I wanted to annoy my children.\\ 
EST & I wanted to cook the collards in order for me to annoy my child.\\
N & I wanted to cook the collards.\\
\hline \hline
protest & The protesters apparently started their protest at the Capitol Building then moved to downtown. We happened to be standing at the corner of 16th and Stout when somebody said that the Police were getting ready to tear-gas a group of demonstrators. We looked around the corner and there were Police everywhere. \\ \hline 
orig & They had blockaded the whole street, and shut down the light rail. \\
Sch & A person said that the group of protesters had protested in a street and in order to block the street. \\
soSN & The protesters wanted to block the street, so the person said for the protesters to protest in the street in order to block it.\\
becauseNS & The person said for the protesters to protest in the street in order to block it because the protesters wanted to block the street.\\
becauseSN & Because the protesters wanted to block the street, the person said for the protesters to protest in the street in order to block it.\\
NS & The person said for the protesters to protest in the street in order to block it. The protesters wanted to block the street.\\ 
EST & The person said for the protesters to protest in the street in order for the protesters to block the street.\\
N & The person said for the protesters to protest in the street in order to block it.\\
\hline
\end{tabular}
\caption{Additional Examples of EST outputs \label{story-table}}
\end{small}
\end{table*}

\vspace{-0.25in}

\bibliographystyle{acl}

\end{document}